\documentclass[runningheads]{llncs}
\usepackage{graphicx}
\usepackage{amsmath}
\usepackage{amssymb}
\usepackage{multicol}
\usepackage{multirow}
\usepackage{subcaption}
\usepackage{graphicx}
\usepackage{wrapfig}
\usepackage[export]{adjustbox}

\begin{document}
\title{Self-supervised Landmark Learning with Deformation Reconstruction and Cross-subject Consistency Objectives}
\titlerunning{Landmark Learning with Deformation and Consistency Objectives}
%
\author{Chun-Hung Chao\inst{1} \and
Marc Niethammer\inst{1} 
}
\authorrunning{C.-H. Chao et al.}
%
\institute{University of North Carolina at Chapel Hill 
}
\maketitle              
\begin{abstract}

A Point Distribution Model (PDM) is the basis of a Statistical Shape Model (SSM) that relies on a set of landmark points to represent a shape and characterize the shape variation. In this work, we present a self-supervised approach to extract landmark points from a given registration model for the PDMs. Based on the assumption that the landmarks are the points that have the most influence on registration, existing works learn a point-based registration model with a small number of points to estimate the landmark points that influence the deformation the most. However, such approaches assume that the deformation can be captured by point-based registration and quality landmarks can be learned solely with the deformation capturing objective. We argue that data with complicated deformations can not easily be modeled with point-based registration when only a limited number of points is used to extract influential landmark points. Further, landmark consistency is not assured in existing approaches In contrast, we propose to extract landmarks based on a  given registration model, which is tailored for the target data, so we can obtain more accurate correspondences. Secondly, to establish the anatomical consistency of the predicted landmarks, we introduce a landmark discovery loss to explicitly encourage the model to predict the landmarks that are anatomically consistent across subjects. We conduct experiments on an osteoarthritis progression prediction task and show our method outperforms existing image-based and point-based approaches.

\end{abstract}
\section{Introduction}
\label{sec:intro}
A Point Distribution Model (PDM) is an important approach to describe shape variations for biomedical applications \cite{gardner2013point,gerig2001shape,harris2013statistical}. PDMs rely on a set of landmark points (hereafter "landmarks") that usually correspond to certain anatomical structures to represent the geometry of each shape instance. As designing landmarks for unseen data or tasks and annotating them is labor intensive, many works have been proposed to automatically identify or place the landmarks \cite{hill2000framework,bhalodia2018deepssm,bhalodia2020self}.

\begin{figure}[!ht]
\centering
\begin{tabular}{cc}
Source Image & Target Image \\
\includegraphics[width=.3\linewidth, valign=m]{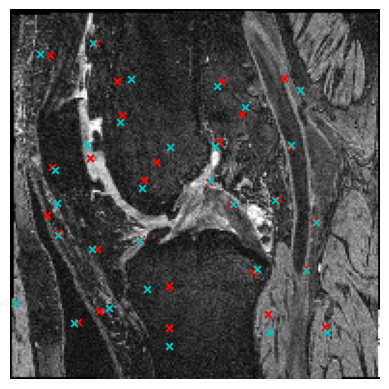} & \includegraphics[width=.3\linewidth, valign=m]{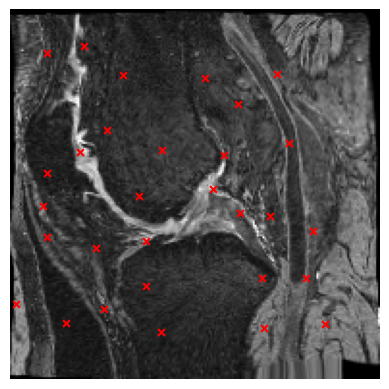} \\[3.3em]
\end{tabular}
\caption{Toy example using the approach by Bhalodia et al. \cite{bhalodia2020self} to find landmarks on a synthesized image pair. Red crosses on the target image indicate the found landmarks; red crosses on the source image indicate the found corresponding landmarks on the source image; and the cyan crosses denote the gold-standard correspondences for the found landmarks on the target image.}
\label{fig:toy}%
\end{figure}

To extract the landmarks for PDMs, conventional approaches require prior knowledge of the shape-of-interest or point-of-interest and rely on segmentation, meshing, or other pre-processing in addition to the image itself to provide guidance for landmark localization. Cates et al. \cite{cates2007shape} proposed a way to extract landmarks by optimizing a set of surface points with the objective of simultaneously maximizing geometric accuracy and statistical simplicity. Similarly, for left atrium modeling in ShapeWorks \cite{cates2017shapeworks}, the binary segmentation of the shape-of-interest is also needed to find landmarks. Agrawal et al. \cite{agrawal2021learning} proposed a deep-learning approach based on a supervised Siamese training paradigm which is the basis of existing landmarks localization approaches \cite{cates2007shape,cates2017shapeworks}. To minimize pre-processing steps, such as region-of-interest delineation, DeepSSM \cite{bhalodia2018deepssm} directly trains a network to predict landmarks under a supervised learning setting with the labels coming from conventional PDM landmark localization methods. As the above approaches require additional labels for supervised learning, the recent work by Bhalodia et al. \cite{bhalodia2020self} presented a self-supervised learning approach to regress landmarks by learning a point-based image registration model. This method assumes that its point-based parametric registration using only a  limited number of points can lead to accurate registration. Therefore it can not capture complex deformations.
Using a toy example with this approach~\cite{bhalodia2020self} (Fig. \ref{fig:toy}) to find 30 corresponding control point pairs to register a 2D image pair that is deformed by a synthesized stationary velocity field, we observe that a correspondence error exists among the found optimal point pairs for registration.  Hence, without making assumptions of which registration model is the best fit for the data, we instead treat the landmark estimation in a knowledge distillation manner and extract the points that are most crucial to the deformation and their correspondences based on a \emph{given} registration model.

\vspace{-1em}

\vspace{-1.5em}

Moreover, the discovered landmarks should not only be important to the deformation, but should also consistently appear on anatomically corresponding locations across the image population, i.e., they should exhibit anatomical consistency. Thus, we design a landmark discovery loss that minimizes the distance between the same landmarks from a triplet of images after mapping them to a common coordinate to enforce the consistency of our landmark predictions. Additionally, since there are a lot of different deforming parts in the data with complex deformations, the model needs to generate a sufficient number of landmarks to be able to select the best points to describe the deformation. As traditional point proposal architectures, e.g., based on a multilayer perceptron (MLP) \cite{bhalodia2020self,bhalodia2018deepssm} or a heatmap \cite{nibali2018numerical,ma2020volumetric}, require a significant amount of memory when adopted for applications in 3D medical images, we propose to use a grid-based architecture to reduce computational resource consumption.

\textbf{Contributions}: (1) We design a framework that extracts landmarks in a registration-model-agnostic manner, so it can be applied to data that requires state-of-the-art registrations rather than using simple point-based registration models with a limited number of control points. (2) We introduce a landmark discovery loss that enforces anatomical consistency between the landmarks. (3) We demonstrate the effectiveness of our framework in proposing landmarks to predict osteoarthritis disease progression based on the challenging Osteoarthritis Initiative (OAI) data set. 

\section{Methodology}
\label{sec:method}

\subsection{Overview}
\begin{figure}
    \begin{center}
        \includegraphics[width=0.7\linewidth]{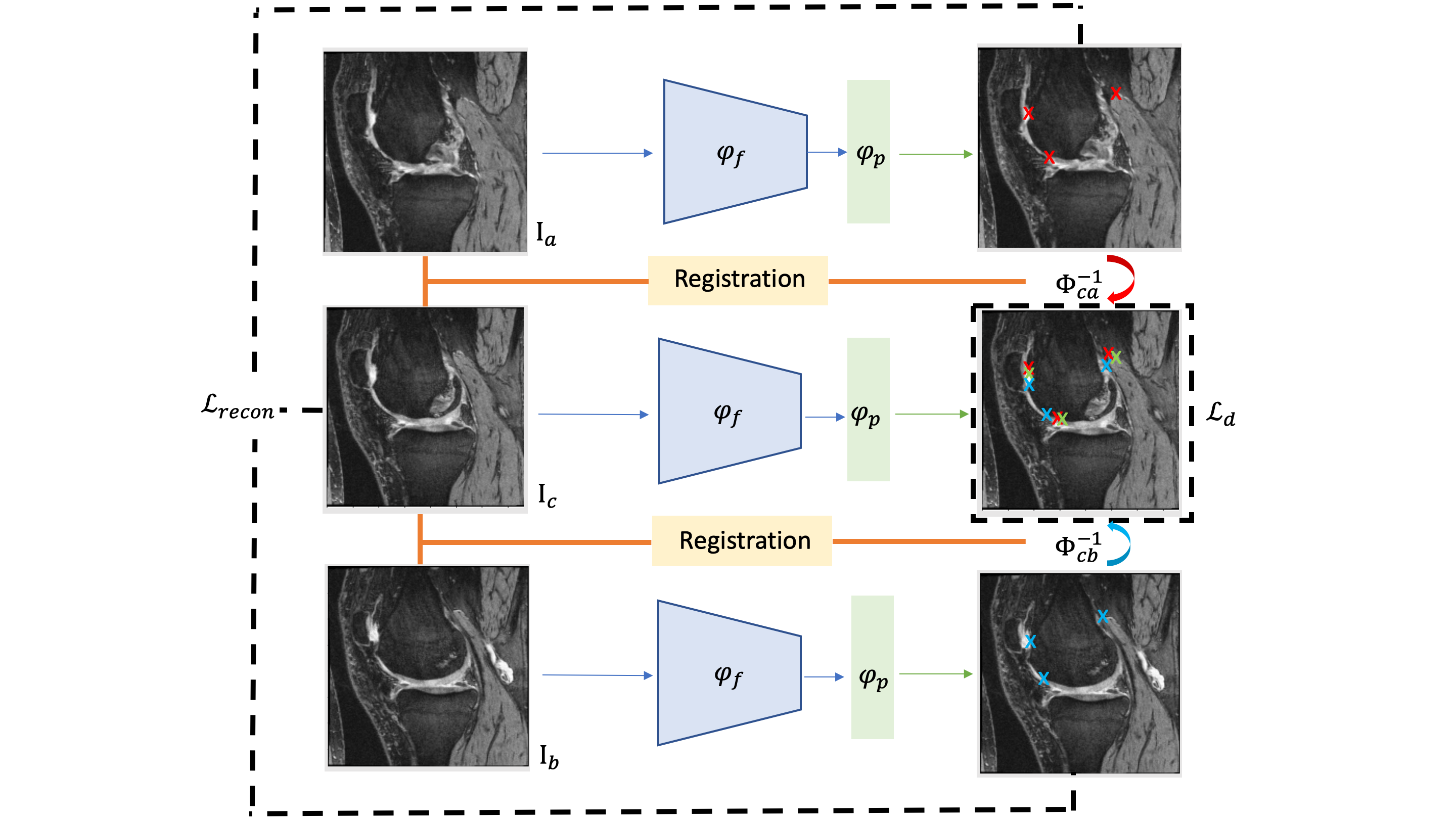}
    \end{center}
  \caption{Overview of our framework. Our framework takes three images $I_a$, $I_b$, and $I_c$ as input. Through feature extractor $\psi_f$ and point-proposal head $\psi_p$, our framework predicts a set of landmarks on each image. Leveraging the transformation field $\Phi^{-1}$, we compute the deformation reconstruction loss $\mathcal{L}_{recon}$ and map all the landmarks to $I_c$ to compute the landmark discovery loss $\mathcal{L}_d$. }
  \label{fig:overview}
\end{figure}

Given three images $\{I_a, I_b, I_c\}$ sampled from a set of training images, we feed them separately into our point proposal network to obtain three sets of landmark predictions, $\{P_a, P_b, P_c\}$. Along with the landmark predictions, we also perform image registration on the image pairs $(I_c, I_a)$ and $(I_c, I_b)$ with the given registration model and obtain transformation fields $\Phi^{-1}_{ca}$ and $\Phi^{-1}_{cb}$. Based on the landmark predictions and the transformation fields, we compute the deformation reconstruction loss $\mathcal{L}_{recon}$ and landmark discovery loss $\mathcal{L}_d$ to train our point proposal network. Fig. \ref{fig:overview} gives an overview of our framework.

\subsection{Point Proposal Network}
Our point proposal network consists of two modules: (1) a convolutional neural network (CNN) that serves as the image feature extractor $\psi_f$ and (2) a point proposal head $\psi_p$ that predicts landmark coordinates from the extracted image features. Given a 3D image $I$ of size $D \times H \times W$, the image feature extractor $\psi_f$ outputs features $f \in \mathbb{R}^{(d \times h \times w) \times c}$ of image $I$. Let $G_i \in \mathbb{R}^3$ denote the coordinate of the $i$-th grid point on a regular grid with $N = d \times h \times w$ grid points and uniform spacing $\delta$ such that $[D, H, W]^{T} = \delta [d, h, w]^{T}$. Our grid-based point proposal head $\psi_p: \mathbb{R}^c \rightarrow \mathbb{R}^3$ predicts a displacement vector for each grid point to produce the coordinate of the $i$-th landmark, $p_i$:
\begin{align}
    p_i &= \psi_{p}(f_i) + G_i\,, \\
    f &= \psi_{f}(I)\,,
\end{align}
where $f = \{f_1, f_2,\cdots, f_n\}$ denotes all predicted features at $n$ different locations, $f_i$ is the i-th feature, 
and all the landmark predictions of image $I$ are:
\begin{align}
    P=\{p_i | i \in \mathbb{Z}^+, 1 \leq i \leq N\}\,.
\end{align}

\subsection{Landmark Discovery Loss}
\begin{figure}
    \begin{center}
        \includegraphics[width=0.9\linewidth]{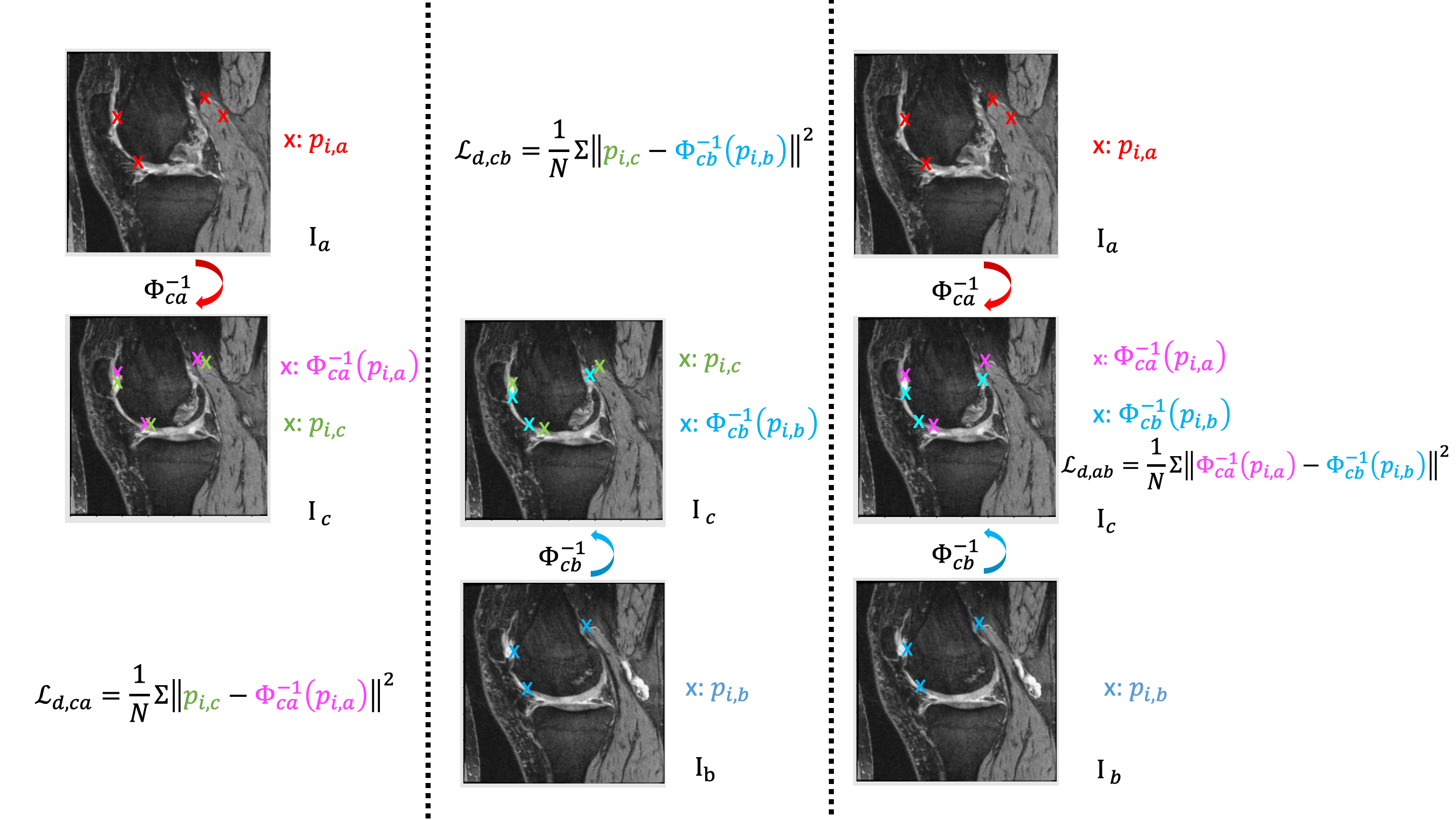}
    \end{center}
  \caption{Graphical Illustration of landmark discovery loss. The red crosses, green crosses, and blue crosses are the detected landmarks on image $I_a$, $I_b$, and $I_c$. With given transformation field $\Phi^{-1}$, we can map the red crosses and the blue crosses to the coordinate system of $I_b$, where we denote the mapped landmarks with the magenta and cyan crosses and compute landmark discovery loss terms $\mathcal{L}_{d, ab}$, $\mathcal{L}_{d, ca}$, and $\mathcal{L}_{d, cb}$. }
  \label{fig:ld_explain}
\end{figure}
\vspace{-1em}
Based on the assumption that a landmark point is a point that can be consistently identified on most images in a population of images, we propose a triplet loss to encourage the point proposal network to discover such consistent landmark points. In other words, we want the landmarks proposed by our model to be anatomically corresponding across images. Achieving the landmark consistency goal requires accurate correspondences of each landmark point between images. Although, in the most ideal scenario, accurate landmark correspondences can be learned simultaneously in a multi-task learning setting where our model jointly learns to perform image registration and landmark prediction, practically this multi-task learning setting often leads to inexact landmark correspondences due to the limited expressive capabilities of landmark-based registration when only a small number of landmarks are extracted. Hence, we advocate decoupling the registration task from the landmark prediction task and leverage provided image registration models to obtain landmark correspondences between images. As there is no constraint on the choice for the registration model as long as it can provide a high quality  transformation field for a pair of images, we can choose the model that is best for our data, e.g., optimization-based fluid-based registration models or deep-learning multi-step registration models that normally have better registration performance \cite{shen2019networks,tian2022gradicon}.

Let $\psi_{reg}$ denote the given registration model from which we obtain transformation fields $\Phi^{-1}_{ca}=\psi_{reg}(I_c, I_a)$ and $\Phi^{-1}_{cb}=\psi_{reg}(I_c, I_b)$, where $\Phi^{-1}_{ca}$ is the map that warps image $I_a$ to image $I_c$ defined in the coordinate system of $I_c$. For the landmark predictions of $I_a$, $P_a$, and the landmark predictions of $I_b$, $P_b$, if they are anatomically equivalent, their corresponding points on $I_c$ should be close to each other. From this idea, we can define our loss:
\begin{align}
    \mathcal{L}_{d, ab} = \frac{1}{N}\sum_{i=1}^{N} \| \Phi^{-1}_{ca}(p_{i, a}) - \Phi^{-1}_{cb}(p_{i, b}) \|^2,
\end{align}
where $p_{i, a}$ is the $i$-th landmark in $P_a$ and $p_{i, b}$ is the $i$-th landmark in $P_b$. To further exploit the input image triplet and the two transformation fields, $\Phi^{-1}_{ca}$ and $\Phi^{-1}_{cb}$, we can use $P_c$ to define two extra terms:
\begin{align}
    \mathcal{L}_{d, ca} = \frac{1}{N}\sum_{i=1}^{N} \| p_{i, c} - \Phi^{-1}_{ca}(p_{i, a}) \|^2\,, \\
    \mathcal{L}_{d, cb} = \frac{1}{N}\sum_{i=1}^{N} \| p_{i, c} - \Phi^{-1}_{cb}(p_{i, b}) \|^2\,,
\end{align}
where $p_{i, c}$ is the $i$-th landmark in $P_c$. Then we define our complete landmark discovery loss as:
\begin{align}
    \mathcal{L}_{d} = \mathcal{L}_{d, ab} + \mathcal{L}_{d, ca} + \mathcal{L}_{d, cb}\,.
\end{align}
Fig. \ref{fig:ld_explain} graphically illustrates each term in the landmark discovery loss.

Although we may also be able to encourage one-directional landmark consistency with one image pair and one transformation field by using the loss
\begin{align}
    \frac{1}{N}\sum_{i=1}^{N} \| p_{i, a} - \Phi^{-1}_{ab}(p_{i, b}) \|^2,
\end{align}
we hypothesize that using a triplet of images and our formulation of $\mathcal{L}_{d}$ is more symmetric and provides more regularity with an acceptable trade-off of performing image registration on one extra pair of images.
\vspace{-0.5em}
\subsection{Deformation Reconstruction Loss}
In addition to landmark consistency, another important attribute for the landmarks is that they are placed on anatomical locations that are able to characterize the deformation as well as possible and can thereby serve as a good structure representation for the PDM. In fact, if we solely rely on the landmark discovery loss $\mathcal{L}_d$ as our training loss, the point proposal network may degenerate to placing all the landmarks on a single location since it can easily minimize $\mathcal{L}_d$ in this manner. To ensure the quality of landmark distribution and motivate the point proposal network to find the points that are crucial towards characterizing the deformation, we introduce a deformation reconstruction loss that penalizes the difference between the original deformation field and the deformation field reconstructed from the landmarks. When reconstructing a dense deformation field from sparse points, a vital design decision to make is the choice of the interpolation method. Common choices in biomedical image registration, like B-splines \cite{rueckert1999nonrigid} or thin plate splines \cite{duchon1977splines}, do not fit well for our task since the former only work for regular grid points, which are not placed according to anatomical information, and the latter involves solving the inverse of a $(N+4) \times (N+4)$ matrix and becomes computationally costly when we have more landmarks, i.e., a larger $N$. Thus, we propose using Nadaraya-Watson interpolation:
\begin{align}
    \Tilde{D}^{-1}_{ab}(\textbf{x}) = \frac{\sum_i K(\textbf{x}, \textbf{p}_{i,b})(\textbf{p}_{i,a} - \textbf{p}_{i,b})}{\sum_i K(\textbf{x}, \textbf{p}_{i,b})}\,,
\end{align}
where $\Tilde{D}^{-1}_{ab}$ is the reconstructed deformation field, $\textbf{x}$ is the location where we want to evaluate the displacement field in the $I_b$ space, and $K(\cdot, \cdot)$ is the kernel function:
\begin{align}
    K(\textbf{x}, \textbf{p}) = e^{-\frac{\|\textbf{x} - \textbf{p}\|^2}{2 \sigma^2}}\,.
\end{align}
The reconstructed transformation field $\Tilde{\Phi}^{-1}_{ab}$ can then be obtained by adding the identity map $\textrm{Id}$ to the reconstructed displacement field:
\begin{align}
    \Tilde{\Phi}^{-1}_{ab} = \Tilde{D}^{-1}_{ab} + \textrm{Id}\,.
\end{align}
With the reconstructed transformation field, we should be able to obtain a warped source image that is similar to the target image:
\begin{align}
    I_a \circ \Tilde{\Phi}^{-1}_{ab} \approx I_a \circ \Phi^{-1}_{ab} \approx I_b \,.
\end{align}
Thus, we define the deformation reconstruction loss for the image triplet as:
\begin{align}
    \mathcal{L}_{recon} = MSE(I_a \circ \Tilde{\Phi}^{-1}_{ac}, I_c) + MSE(I_b \circ \Tilde{\Phi}^{-1}_{bc}, I_c)
\end{align}
where $MSE(\cdot)$ is the mean squared error. Although we can compute the difference with $MSE(\Tilde{\Phi}^{-1}_{ac}, \Phi^{-1}_{ac})$, using the image similarity loss reflects the alignment quality between anatomical structure better and directly minimizing the mean squared error between $\Tilde{\Phi}^{-1}_{ac}$ and $\Phi^{-1}_{ac}$ does not guarantee the alignment of the foreground.

\subsection{Complete Training Loss}
We train our model by combining the landmark discovery loss and the deformation reconstruction loss:
\begin{align}
    \mathcal{L} = \lambda_d \mathcal{L}_d + \lambda_{recon} \mathcal{L}_{recon}\,,
\end{align}
where $\lambda_d>0$ and $\lambda_{recon}>0$ are the hyper-parameters which denote the weight of each individual loss. 

\section{Experimental Results}
\label{sec:result}

\subsection{Experimental Setting}

\textbf{Implementation Details.} We modified DenseNet121 \cite{huang2017densely} as our backbone for the feature extractor in the point proposal network. The feature extractor outputs a feature map corresponding to a regular grid of size $10 \times 24 \times 24$, which gives us $5,760$ landmark predictions for each image. For the external registration model, we use the 4 stages of GradICON \cite{tian2022gradicon} to obtain the transformation fields. We fixed our $\sigma$ in the kernel function of Nadaraya-Watson interpolation at $3$mm. $\lambda_d$ and $\lambda_{recon}$ are set to be $0.005$ and $0.05$ respectively. The network was trained with an ADAM optimizer for $35$ epochs, with the learning rate set to $0.001$ and a learning rate scheduler that decreases the learning rate by $1\%$ after each epoch. The batch size was set to $4$ throughout the training. We implemented the entire network and the Nadaraya-Wastson interpolation module with PyTorch and PyKeops \cite{charlier2021kernel}. The training was done on one NVIDIA RTX A6000 GPU.

\noindent\textbf{Dataset.} We build our OA progression dataset based on the train-val-test patient split of the OA registration dataset in AVSM \cite{shen2019networks}. After filtering out the patients without progression labels, we obtain 474 subjects for training, 20 for validation, and 63 for testing. We determine OA progression based on the often-used Kellgren and Lawrence grading system (KLG) which classifies the OA severity in clinical practice by checking if the KLG score increases. Among the 63 subjects in the testing set, 39 subjects are non-progressive and the rest are progressive.

\noindent\textbf{Evaluation Metrics.}
For the classification tasks in our experiments, we report the accuracy and the average precision (AP) scores. While not every baseline predicts an ordered landmark set, we use two metrics, one for both ordered and non-ordered landmark sets and another one solely for ordered landmark sets, to evaluate the landmark consistency. For both consistency metrics, we first randomly sample a third image from a different subject for each pair of input images and map the detected landmarks on both input images in each pair to the sampled third image using an oracle registration model that provides an oracle transformation map $\Phi^{-1}$. As the two landmark sets are mapped to the third image and form two point clouds in a common coordinate system, we use the mean symmetric Chamfer distance $d_{CD}$ of the two landmark sets:
\begin{align*}
    \frac{1}{2N} (\sum_{i=1}^{N} \min_{j=1,...,N} \| \Phi^{-1}_{ca}(p_{i,a}) - \Phi^{-1}_{cb}(p_{j,b}) \|_2 + \sum_{i=1}^{N} \min_{j=1,...,N} \| \Phi^{-1}_{cb}(p_{i,b}) - \Phi^{-1}_{ca}(p_{j,a}) \|_2)
\end{align*}
for the ordered and non-ordered landmark predictions. Specifically for the ordered landmark sets, we can further evaluate the consistency error using
\begin{align*}
    \frac{1}{N} \sum_{i=1}^N \| \Phi^{-1}_{ca}(p_{i,a}) - \Phi^{-1}_{cb}(p_{i,b}) \|_2\,.
\end{align*}

\subsection{OA Progression Prediction}
\vspace{-3em}
\begin{table}[ht]
\caption{Quantitative results of OA progression prediction.}
\label{tab: quan_oa_prog}
\begin{center}
\begin{tabular}{||cc||c|c|c|c|c||}
    \hline
    \hline
    & & DenseNet & GraphRegNet & Bhalodia et al. & Ours w/o $\mathcal{L}_d$ & Ours \\
    \hline
    \hline
    \multicolumn{2}{||c||}{Acc} & 65.08\% & 58.73\% & 61.90\% & 60.32\% & 68.25\% \\
    \multicolumn{2}{||c||}{AP} & 54.10\% & 37.55\% & 37.86\% & 43.28\% & 61.00\% \\
    \hline
    \multicolumn{2}{||c||}{Chamfer Dist. (mm)} & - & $7.38 \pm 1.77$ & $7.71 \pm 2.45$ & $4.71 \pm 0.46$ & $2.41 \pm 0.47$ \\
    \hline
    \multirow{4}{*}{\shortstack[c]{Ordered \\ Consist. \\ Error \\ (mm)}} 
    & X & - & - & $2.46 \pm 0.76$ & $3.57 \pm 1.56$ & $1.11 \pm 0.33$ \\
    & Y & - & - & $3.21 \pm 1.37$ & $3.48 \pm 1.27$ & $0.95 \pm 0.35$ \\
    & Z & - & - & $2.09 \pm 0.84$ & $3.42 \pm 0.90$ & $1.03 \pm 0.33$ \\
    & All & - & - & $5.33 \pm 1.76$ & $7.10 \pm 1.99$ & $2.14 \pm 0.65$ \\
    \hline
    \hline
\end{tabular}
\end{center}
\vspace{-2em}
\end{table}

Osteoarthritis (OA) is a joint disease that develops slowly over the years. To study the effectiveness of the proposed landmarks, we choose knee OA progression prediction as our downstream task. Given the first two knee Magnetic Resonance (MR) scans separated by 12 months, our goal is to correctly predict if OA will get worse within the next 72 months. For the OA progression prediction task, our model first detects two sets of estimated landmarks on the two initial scans. Then we combine and process the landmarks across patients, which are ordered, by generalized Procrustes analysis and train a linear distance-weighted discrimination (DWD) classification model with the proposed points from the training set to predict whether OA progression will occur.

In order to compare the results, we use one image-based and two point-based methods as our baselines. For the image-based baseline, we train a 3D DenseNet121 to predict the progression directly with the two given scans using a binary cross entropy loss. The first point-based baseline is GraphRegNet \cite{hansen2021graphregnet}, which performs registration using the handcrafted Forstner keypoints. We combine the detected F\"orstner keypoints on the second input scan and their corresponding points on the first input scan, which are obtained with the transformation field estimated by GraphRegNet, to form the point representation. The second point-based baseline is the work done by Bhalodia et al. \cite{bhalodia2020self}. Bhalodia et al. proposed to train a network to predict landmarks on the two input scans by learning a point-based registration model with regularization on the coefficients of the thin-plate spline interpolation. Due to the memory intensive nature of the work by Bhalodia et al., this baseline only predicts $152$ landmarks so it can fit in the GPU we used. We process the point representation of the baselines using the same protocol we used to process the landmarks for our approach. Ordered landmarks are processed with generalized Procrustes analysis.

Tab. \ref{tab: quan_oa_prog} shows the quantitative OA progression prediction results. Training a DenseNet classifier directly from images yields $65.08 \%$  in accuracy and $54.10 \%$ in average precision and serves as a strong baseline for OA progression prediction. Our method produces $68.25 \%$  in accuracy and $61.00 \%$ in average precision. This shows that our method successfully extracts the disease-related information and is an efficient representation, as each input image is characterized using $5,760$ landmarks. With handcrafted F\"orstner keypoints, GraphRegNet produces the lowest performance as the keypoints are originally designed for lung CT and might not be suitable for knee MR. Due to memory constraints, the approach by Bhalodia et al. can only predict $152$ landmarks and yields an average precision score of only $37.86\%$, which is possibly due to the fact that the number of landmarks insufficient to characterize images pairs with complex deformations. In fact, for our method with merely $152$ landmarks, it could only reach AP of $38.4\%$. We observe that the Chamfer landmark distance consistency evaluation is correlated with the average precision score. In fact, by comparing the results between our method with and without $\mathcal{L}_d$, we can conclude it is advantageous to enforce landmark consistency when building the point representation.

\subsection{Qualitative Results}

\begin{figure}[!ht]
\centering
\begin{tabular}{ccccc}
& Landmark 1 & Landmark 2 & Landmark 3 & Landmark 4\\
& (0.1166) & (0.1159) & (0.1058) & (0.1009) \\[0.5em] 
Patient 1 & 
\includegraphics[width=.18\linewidth, valign=m]{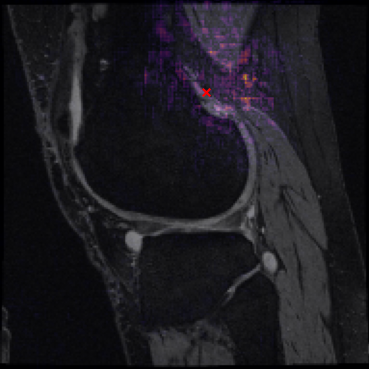} & \includegraphics[width=.18\linewidth, valign=m]{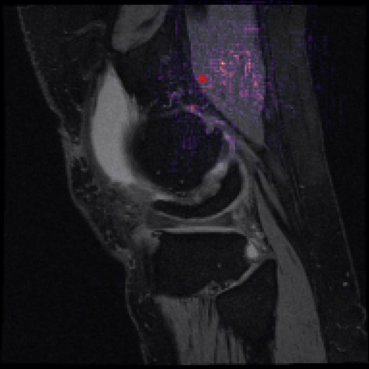} & \includegraphics[width=.18\linewidth, valign=m]{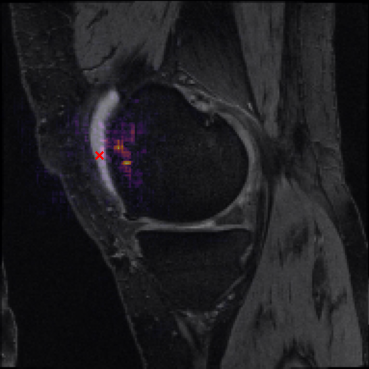} &
\includegraphics[width=.18\linewidth, valign=m]{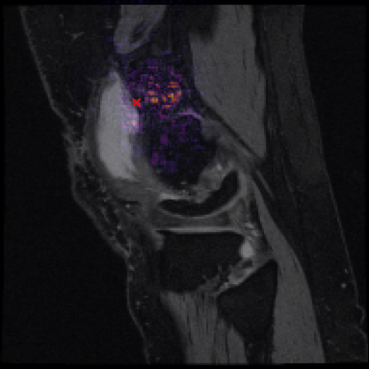} \\[2.7em]
Patient 2 & 
\includegraphics[width=.18\linewidth, valign=m]{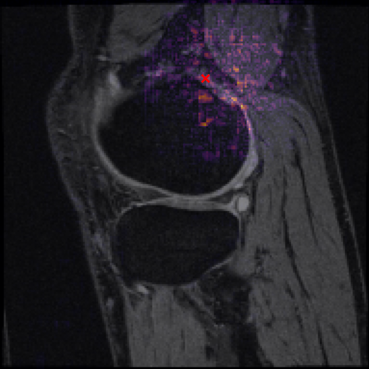} & \includegraphics[width=.18\linewidth, valign=m]{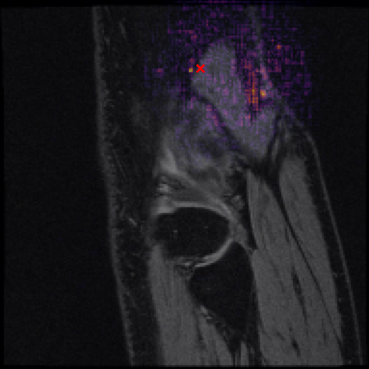} & \includegraphics[width=.18\linewidth, valign=m]{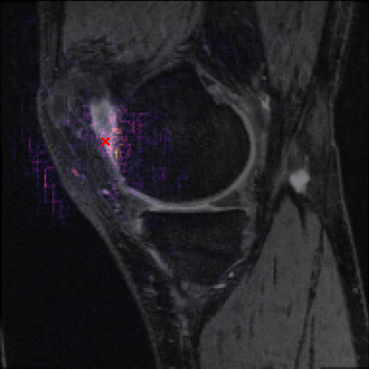} &
\includegraphics[width=.18\linewidth, valign=m]{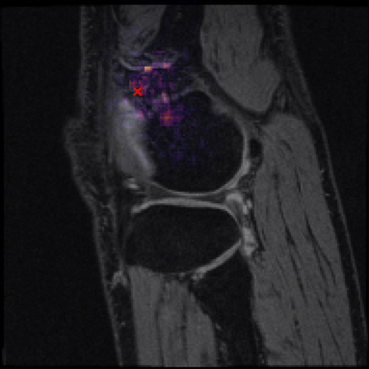} \\[2.7em]
Patient 3 & 
\includegraphics[width=.18\linewidth, valign=m]{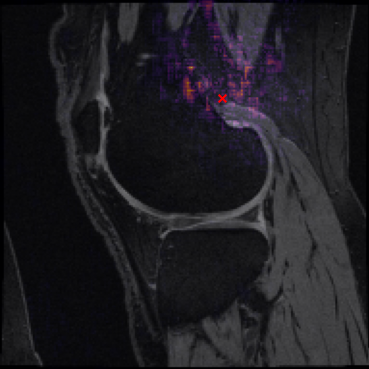} & \includegraphics[width=.18\linewidth, valign=m]{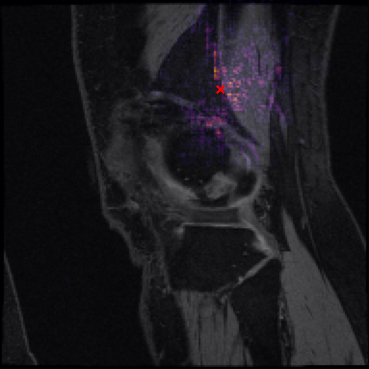} & \includegraphics[width=.18\linewidth, valign=m]{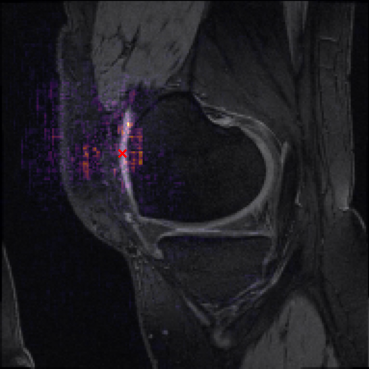} &
\includegraphics[width=.18\linewidth, valign=m]{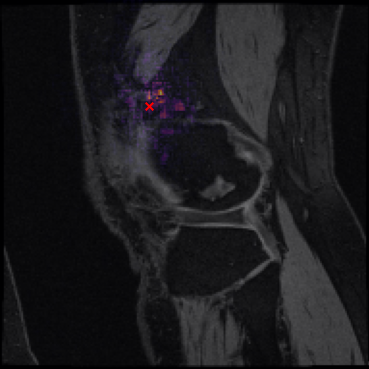}
\end{tabular}
\caption{Qualitative results of landmark predictions. The intensity of the purple indicates the magnitude of the gradient activation.  The red crosses indicate the landmarks. The numbers in the parentheses are the corresponding DWD coefficient magnitudes.}
\label{fig:qual}%
\vspace{-2em}
\end{figure}

In this section, we visualize the top influential landmarks and their gradient activation map to provide visual interpretations for the landmark predictions.

\noindent\textbf{Gradient Activation Map.} We adopt a strategy similar to GradCAM \cite{selvaraju2017grad} by computing the gradient activation using the magnitude of $\frac{\partial \| \psi_p(f_i) \|^2}{\partial I(x)}$ to highlight the voxels that have the greatest impact on the landmark predictions. From Fig. \ref{fig:qual}, we can see that the gradient responses primarily occur on local structures near the landmarks.

\begin{wrapfigure}{r}{5cm}
    \vspace{-3em}
    \includegraphics[width=5cm]{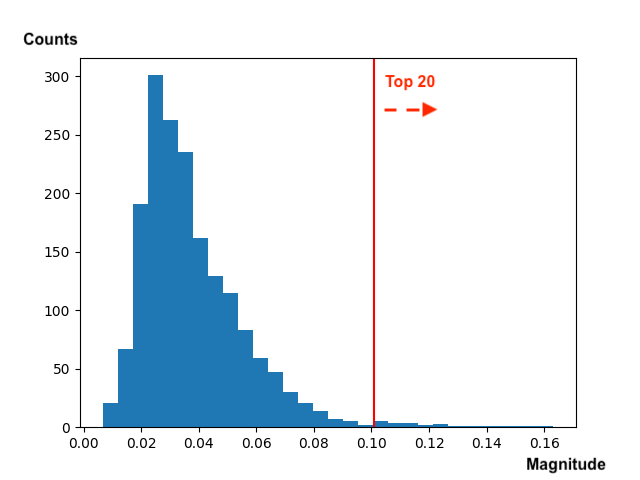}
    \caption{Histogram of the DWD coefficient weights and occurrence frequencies. The few but important (larger weight) landmarks are ideal image markers for OA progression prediction. }
    \label{fig:dwd_coef_hist}
    \vspace{-0.5em}
\end{wrapfigure}

\noindent\textbf{Top Influential Landmarks.} As we use linear DWD for progression prediction, we can quantify the importance of each landmark by comparing the sum of the weights operating on each landmark value, which indicates the landmark influence for identifying a positive case. Fig. \ref{fig:dwd_coef_hist} shows the weight distribution of our trained DWD model and Fig. \ref{fig:qual} shows the landmarks randomly sampled from the top twenty influential landmarks from different patients. We can observe that the predicted landmarks are anatomically consistent across different patients. Critically, since these landmarks are important factors for the linear DWD model, these landmarks have the potential to serve as image markers for OA progression prediction. \\

\begin{wraptable}{r}{4cm}
\caption{Relationship between the number of landmarks and the average precision of classification}
\label{tab: n_landmarks_vs_ap}
\vspace{-1em}
\begin{tabular}{cc}\\
\hline
N of landmarks & AP ($\%$) \\
50 & 51.8 \\
150 & 56.3\\
152 & 57.2 \\
250 & 55.6 \\
550 & 57.9 \\
1250 & 61.0 \\
\hline
\end{tabular}
\vspace{-1em}
\end{wraptable} 

In addition, we discuss the relationship between the number of landmarks and the classification performance (shown in Tab. \ref{tab: n_landmarks_vs_ap}). We trained DWD using different numbers of top influential landmarks. Using the top $50$ landmarks yields the average precision value of $51.8\%$, and increasing the number of landmarks generally has a positive effect on the classification performance. It is worth noticing that when using $152$ landmarks (same number as Bhalodia et al.), our method outperforms Bhalodia et al. with a large margin ($55.6\%$ vs $37.9\%$). This shows that our method can produce more meaningful landmarks even with a limited number of landmarks.

\vspace{-1em}

\section{Conclusion}
In this work, we presented a learning-based landmark proposal framework. With our deformation reconstruction and consistency objectives, the model learns to predict the points that not only characterize the deformation well but also are anatomically consistent across patients and constitute landmarks points. In our experiments on OA progression prediction, our method achieves the best prediction performance as well as the lowest consistency error. Based on the quantitative and qualitative results, we demonstrated that our proposed landmarks can serve as point representations of images and can potentially be used to discover prognostic image markers for early disease diagnosis. 

\vspace{-1em}
\section{Acknowledgements}
This work was supported by NIH grants 1R01AR072013 and R41MH118845. The work expresses the views of the authors, not of NIH. The knee imaging data were obtained from the controlled access datasets distributed from the Osteoarthritis Initiative (OAI), a data repository housed within the NIMH Data Archive. OAI is a collaborative informatics system created by NIMH and NIAMS to provide a worldwide resource for biomarker identification, scientific investigation and OA drug development. Dataset identifier: NIMH Data Archive Collection ID: 2343.

\vspace{-1em}

%
%
%
\bibliographystyle{splncs04}
\bibliography{mybibliography}
%




\end{document}